\documentclass{article}
\usepackage{ijcai26}

\usepackage{times}
\usepackage{soul}
\usepackage{url}
\usepackage[hidelinks]{hyperref}
\usepackage[utf8]{inputenc}
\usepackage[small]{caption}
\usepackage{graphicx}
\usepackage{amsmath}
\usepackage{amsthm}
\usepackage{booktabs}
\usepackage{algorithm}
\usepackage[switch]{lineno}
\usepackage{fontawesome5}
\usepackage{amsmath}
\usepackage{amsthm}
\usepackage{algorithm}
\usepackage{algpseudocode}
\usepackage{booktabs}
\usepackage{mathtools,amssymb}
\usepackage{subfigure}
\usepackage{booktabs} 
\usepackage{bm,bbm}
\usepackage{multirow}
\usepackage{wrapfig}
\usepackage{hyperref}       % hyperlinks
\usepackage{url}            % simple URL typesetting
\usepackage{threeparttable}

\usepackage{breqn}
\usepackage{titletoc}
\usepackage{appendix}
\usepackage{etoc}

\usepackage[dvipsnames]{xcolor}
\usepackage{tcolorbox}
\newcommand{\faHuggingFace}{\includegraphics[height=1.1em]{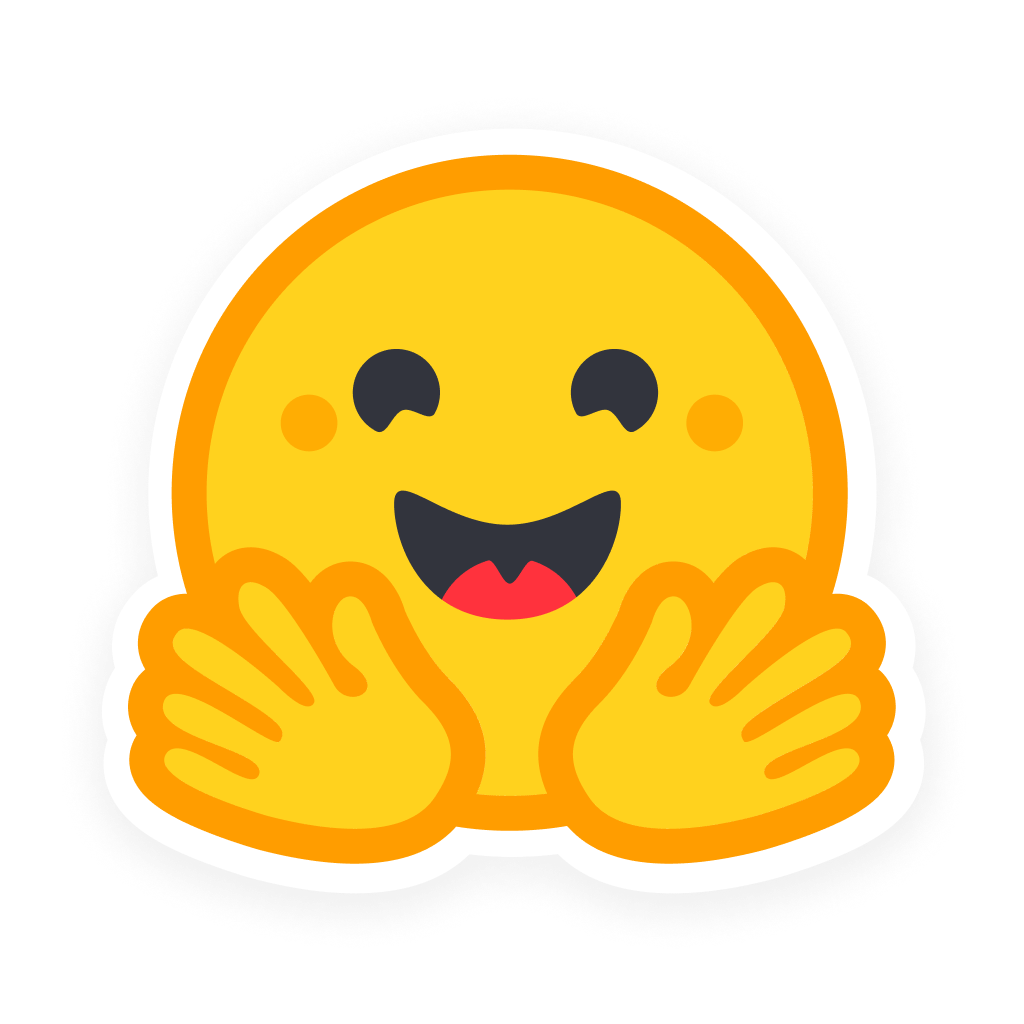}}

\title{Speaking Numbers to LLMs: Multi-Wavelet Number Embeddings for Time Series Forecasting
}

\author{
Defu Cao$^{1}$\footnote{Equal contribution. Correspondence to: Defucao@usc.edu. This work was done while Zijie Lei was at USC.}
\and
Zijie Lei$^{1*,2}$\and
Muyan Weng$^1$\and
Jiao Sun$^{1,3}$\And
Yan Liu$^1$\\
\affiliations
$^1$University of Southern California\\
$^2$ Meta \\
$^3$Google DeepMind\\
\emails
\{defucao, zijielei, muyanwen, jiaosun, yanliu.cs\}@usc.edu
}
\begin{document}

\maketitle

\begin{abstract}
Large language models (LLMs) are attractive for context-aware time series forecasting because they can integrate heterogeneous textual signals, yet their discrete, language-oriented tokenization and embedding interfaces are misaligned with continuous numerical values, often harming numerical ordering and forecasting reliability. We propose TempoWave, a plug-and-play temporal wavelet digit interface that maps each scalar observation into digit-wise embeddings constructed from multi-wavelet, multi-scale coefficients. By directly overriding standard token representations, TempoWave seamlessly exposes both fine-grained local fluctuations and macro global structures in a transformer-compatible form, ensuring that precise numerical formatting, distinct digit identity, and robustness to common normalization operations are maintained throughout the LLM pipeline. Experiments across five context-enriched forecasting benchmarks demonstrate that TempoWave consistently improves LLM-based forecasters over standard numeric tokenization and alternative embedding interfaces, achieving a new state-of-the-art. These results highlight the numeric interface as a key bottleneck and suggest that principled multi-resolution embeddings can better couple LLMs' contextual reasoning with precise forecasting. Our code is available at \href{https://github.com/DC-research/TempoWAVE}{\faGithub~DC-research/TempoWAVE} and our model can be accessed at \href{https://huggingface.co/Melady/TempoWAVE}{\faHuggingFace~Melady/TempoWAVE}.
\end{abstract}

\section{Introduction}
Time series analysis, the study of data points ordered chronologically, is indispensable across sectors such as finance, healthcare, and climate science~\cite{cao2025conversational,wang2026se}.  Accurate forecasting supports resource allocation, risk management, and early warning systems, yet the underlying data generating processes are often complex and evolving~\cite{cao2023estimating,zhang2022counterfactual}.  Practical time series typically exhibit non-stationarity, mixed periodicities, regime shifts, long- and short-range temporal dependencies, and substantial noise~\cite{non-stationary,cao2020spectral,9561461,LLFiance}.  These properties make it difficult to learn models that simultaneously capture fine-grained local fluctuations and long-horizon global structure, while remaining robust under distribution shifts and limited supervision.

In parallel, Large Language Models (LLMs)~\cite{GPT-4} have become strong general-purpose sequence learners.  They can exploit long contexts, perform in-context pattern induction, and naturally integrate textual information~\cite{hu2025contextalignment,zhou2025can,zhang2024guiding}.  This is particularly appealing for time series intelligence because many exogenous drivers that affect temporal dynamics are expressed in language, such as policy changes, market news, clinical narratives, and operational logs.  Moreover, the few-shot and zero-shot generalization behavior of LLMs suggests a promising pathway for domains where labeled time series data is scarce and task distribution varies across entities, locations, or time periods.

Despite this promise, directly adapting LLMs to time series forecasting remains challenging.  LLMs are optimized for discrete token prediction, whereas time series forecasting fundamentally requires precise modeling of continuous values.  This mismatch can lead to unreliable numerical behavior even when the model captures high-level temporal patterns~\cite{merrill2024language,ye2025llm}.  More critically, language-oriented tokenization fragments numbers into sub-tokens in ways that are not tied to magnitude, for example ``2026'' $\rightarrow$ ``20'' and ``26''.  Such fragmentation breaks ordinal relations and obscures the continuity intrinsic to temporal processes.  As a result, two numerically close values may be mapped to very different token sequences, while numerically distant values can share sub-tokens, introducing spurious similarity.  In LLM-based forecasting pipelines, this translation layer between real-valued sequences and discrete tokens becomes a principal bottleneck.

Recent research has pursued several avenues to bridge the gap between LLMs and time series analysis.  One direction develops specialized foundation models tailored to time series~\cite{cao2026pinfdit,cao2023tempo}.  Another uses agentic or multimodal systems that couple LLMs with dedicated forecasting tools~\cite{ye2024beyond,anonymous2026someone}.  A third direction focuses on input adaptations, including patching~\cite{Yuqietal-2023-PatchTST}, quantization~\cite{talukdertotem}, or converting time series into symbolic or textual representations~\cite{GPT4MTS}.  While these approaches can improve usability and efficiency, they often trade away numerical faithfulness, blur fine-grained variations, or rely on external components that reduce end-to-end differentiability and complicate analysis.  Consequently, there remains a persistent gap in representing continuous numerical values inside the standard transformer input space in a principled and information-preserving manner.

In this work, we focus on the representation bottleneck and ask whether LLMs can be equipped with numerically grounded embeddings that preserve quantitative relations and multi-scale temporal structure while remaining compatible with standard transformer inputs.  An effective forecasting representation should support reasoning across resolutions.  Local changes are crucial for short-term dynamics and anomaly-sensitive regimes, while trends and seasonal components dominate long-horizon behavior.  Motivated by wavelet analysis, which provides a natural multi-resolution decomposition, we propose \textit{Multi-Wavelet Number Embedding (TempoWave)}, an input embedding interface that maps each scalar observation into a dense vector encoding multi-scale structure.  TempoWave is designed to be injected into LLM backbones without requiring language tokenization of numbers, thereby reducing the disconnect between numeric magnitude and the model's discrete interface.
Beyond forecasting accuracy, we also aim to understand how the embedding interface shapes numerical structure inside the model.  To this end, we introduce diagnostic analyses that probe whether local neighborhoods in representation space respect numeric ordering, a property we refer to as {monotonic neighborhood consistency}.  These analyses help explain when TempoWave improves forecasting and provide guidance for designing numerically grounded interfaces for LLMs.  Extensive experiments on diverse forecasting benchmarks show that TempoWave consistently improves LLM-based forecasters compared to standard tokenization and common adaptation methods, and it is competitive with strong time-series-specific models in settings requiring precise numerical forecasting.

\textbf{Contributions.}
Our contributions are as follows:
\begin{itemize}
    \item \textbf{A wavelet-based numeric interface for LLM forecasting.}
    We propose \emph{Multi-Wavelet Number Embedding (TempoWave)}, an input embedding interface that maps each real-valued numerical observation into a dense vector with multi-resolution structure, enabling direct use of standard LLM backbones for numerical forecasting without relying on language tokenization of numbers.

    \item \textbf{Structural Faithful and Stable Numeric Representation.}
    We analyze TempoWave from a multi-scale signal processing perspective and establish properties related to numerical faithfulness and stability, including improved separability across values and robustness to common normalization operations used in transformers.

    \item \textbf{Comprehensive evaluation with diagnostic evidence.}
    We conduct extensive experiments on diverse forecasting benchmarks and show consistent gains over LLM baselines using tokenization and common input adaptations.  We further provide diagnostic analyses that probe {monotonic neighborhood consistency}, offering evidence for how TempoWave reshapes numerical neighborhoods and helping explain its forecasting improvements.
\end{itemize}

\section{Related Work}
\label{sec:related_work}

Research on applying Large Language Models (LLMs) to time series analysis has grown rapidly, motivated by LLMs' strong sequence modeling and their ability to integrate contextual information. Existing efforts can be grouped into three directions: time series foundation models trained on large-scale temporal corpora, LLM-centered agentic or multimodal systems, and input adaptation strategies that re-design how numerical values are represented for transformer backbones.

\paragraph{Time series foundation models.}
A major direction is to pre-train foundation models directly on large and diverse time series collections to learn transferable temporal representations. Representative examples include Chronos~\cite{ansari2024chronos}, TimesFM~\cite{das2024decoder}, and other large-scale temporal pre-training efforts~\cite{woo2024moirai,cao2024timedit,yang2025foundation}. These models demonstrate the benefits of scaling and pre-training for forecasting, but they often rely on patching, discretization, or quantization to interface with transformers, which can blur fine-grained numerical differences and introduce information loss in precision-sensitive regimes.

\paragraph{LLM agents and multimodal time series systems.}
Another line of work uses general-purpose LLMs as reasoning engines within larger pipelines, where forecasting or numerical computation is delegated to specialized tools and the LLM performs orchestration and interpretation~\cite{yang2026adaptive,weng2026temporalbench,yang2025toward}. Closely related are multimodal systems that jointly model time series and text, enabling context-aware forecasting and question answering, for example TimeLLM~\cite{jin2024timellm}, GPT4MTS~\cite{GPT4MTS}. While these approaches highlight the value of fusing textual signals, their performance still depends critically on how continuous values are encoded for transformer inputs, and numerical faithfulness can remain a bottleneck when the interface is token-based or heavily discretized.

\paragraph{Input adaptation and numeric representation for LLMs.}
A substantial body of work focuses on making numerical time series consumable by LLM backbones via input adaptation. Common strategies include patch-based representations~\cite{Yuqietal-2023-PatchTST}, discretization or binning~\cite{talukdertotem}, symbolic conversion of segments into strings~\cite{goswami2024moment}, and embedding alignment methods that map time series embeddings into the language embedding space~\cite{gruver2024large,zeng2023transformers}. More generally, recent studies on numeracy in language models emphasize that tokenization and discrete interfaces can impede faithful numerical reasoning, motivating alternative representations that preserve quantitative structure~\cite{merrill2024language,gillman2025fourier}. These adaptations improve compatibility, but they often shift the burden to a preprocessing stage and may sacrifice either precision, locality, or multi-scale structure~\cite{anonymous2026position}.

\begin{figure*}[ht]
    \centering
    \includegraphics[width=0.9\linewidth]{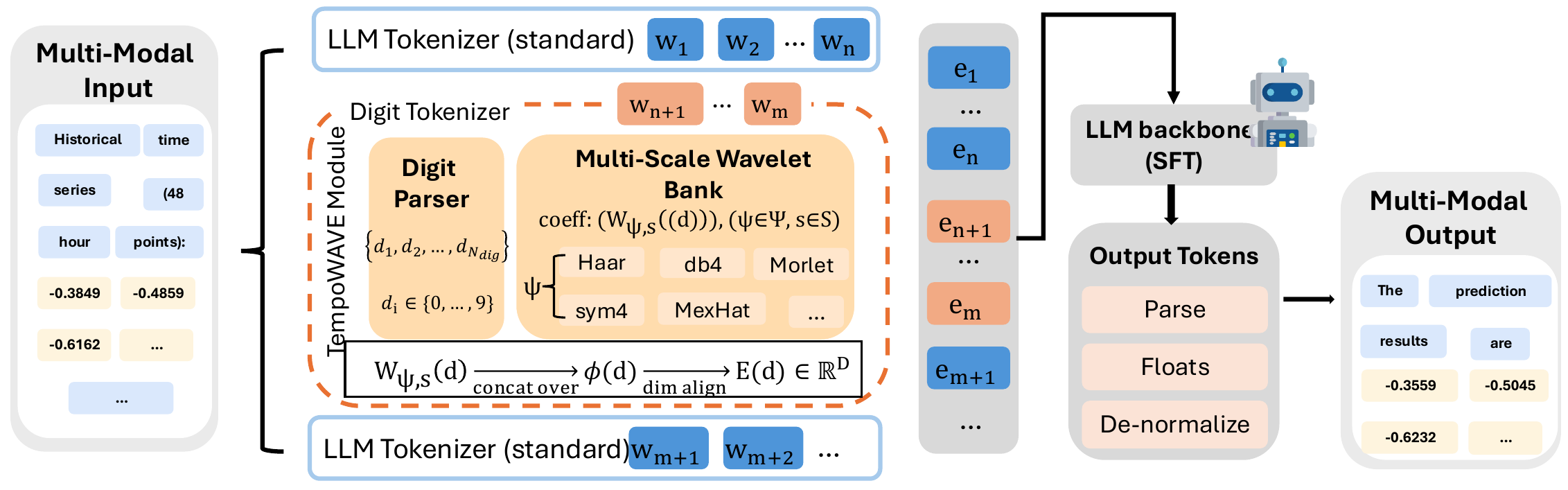}
    \caption{Overview of the TempoWave-based forecasting framework with digit-level tokens.
The input prompt is tokenized once using a tokenizer augmented with dedicated digit tokens.
Text and context tokens use standard embeddings, while digit tokens are routed to the TempoWave module,
which constructs digit embeddings via multi-wavelet, multi-scale coefficients and overrides the corresponding token embeddings.
The resulting embedding sequence is fed into an unchanged LLM backbone trained via supervised fine-tuning (SFT).
The model generates numeric tokens that are parsed, de-normalized, and evaluated as real-valued forecasts.
}
    \label{fig:TempoWave_framework}
\end{figure*}
\paragraph{Positioning of TempoWave.}
Our work addresses the above interface challenge by introducing Multi-Wavelet Number Embedding, which constructs numerically grounded embeddings that encode multi-resolution structure prior to ingestion by an LLM. In contrast to purely symbolic conversion or coarse discretization, TempoWave aims to preserve quantitative relations while providing a multi-scale representation inspired by wavelet analysis, supporting both accurate forecasting and diagnostic analysis of the induced numerical neighborhood structure.

\section{Methodology}
\label{sec:methodology}

\subsection{Overview}
To bridge the mismatch between continuous-valued time series and the discrete input interface of Large Language Models (LLMs), we propose {Multi-Wavelet Number Embedding (TempoWave)}, a numerically grounded embedding interface that intervenes only at the token embedding layer while keeping the LLM backbone unchanged.
As illustrated in Figure~\ref{fig:TempoWave_framework}, TempoWave enables LLMs to process numerical sequences by replacing the embeddings of digit tokens with multi-resolution wavelet-based representations.

\paragraph{Numeric-to-token interface.}
Given a normalized time series value $x_t \in \mathbb{R}$, we first render it into a fixed-precision string with $m_{prec}$ integer digits and $n_{prec}$ fractional digits (e.g., \texttt{V.FFFF}).
A single tokenization pass is then applied using a tokenizer augmented with \emph{dedicated digit tokens}, where each digit $d_i \in \{0,\ldots,9\}$ is treated as an individual token.
As a result, the input prompt is converted into a mixed token sequence consisting of text/context tokens and digit tokens.

Standard token embeddings are used for text and context tokens.
For digit tokens, TempoWave computes digit embeddings via multi-wavelet, multi-scale features and \emph{overrides} the standard embeddings.
This routing mechanism ensures that numerical structure is injected at digit positions only, without altering the remaining token representations.

\paragraph{TempoWave embedding override.}
For each digit token $d$, TempoWave computes a set of wavelet coefficients $W_{\psi,s}(d)$ over a predefined wavelet family $\Psi$ and scale set $S$.
These coefficients are concatenated into a digit feature vector $\phi(d)$ and mapped to the LLM embedding dimension via a fixed alignment function $g(\cdot)$, producing the digit embedding $E(d)\in\mathbb{R}^{D}$.
The resulting digit embeddings replace the standard embeddings at digit positions, yielding the final input embedding sequence
$\mathbf{H}_0 \in \mathbb{R}^{T \times D}$, which is fed into the LLM.

\paragraph{Context and training objective.}
For context-aware forecasting, we fine-tune the LLM via supervised fine-tuning (SFT) on prompts that include
(i) historical numeric values represented as fixed-precision digit tokens,
(ii) optional global descriptors such as Catch22 features~\cite{lubba2019catch22}, and
(iii) situational context such as date and domain information.
The LLM backbone remains unchanged, and training minimizes the standard next-token cross-entropy loss to generate future numeric tokens corresponding to the next $k$ time steps.

\subsection{TempoWave Construction}
\label{sec:TempoWave}

\paragraph{Wavelet dictionary.}
Let $\psi(t)$ denote a mother wavelet and $\psi_{s,\tau}(t)$ denote its scaled and translated version:
\begin{equation}
\psi_{s,\tau}(t) = \frac{1}{\sqrt{s}}\psi\left(\frac{t-\tau}{s}\right),
\label{eq:wavelet_scale_translate_main}
\end{equation}
where $s>0$ is the scale and $\tau$ is the translation. We select a set of wavelets $\Psi=\{\psi_1,\ldots,\psi_k\}$ and a set of scales $S=\{s_1,\ldots,s_l\}$. For TempoWave we use a fixed translation (typically $\tau=0$). 

\paragraph{Digit signal and wavelet coefficients.}
Each digit $d\in\{0,\ldots,9\}$ is normalized to $\tilde d = d/9 \in [0,1]$. To obtain wavelet coefficients without degeneracy from the zero-mean property of admissible wavelets, we represent a digit as a discrete impulse on a fixed grid.
Let $B$ be the grid resolution and $t_r=\frac{r}{B-1}$ for $r=0,\ldots,B-1$. Define the digit index $q(d)=\lfloor \tilde d\,(B-1) \rceil$, and the digit signal $f_d\in\mathbb{R}^{B}$ as a Kronecker delta:
\begin{equation}
(f_d)_r =
\begin{cases}
1, & r=q(d),\\
0, & \text{otherwise}.
\end{cases}
\label{eq:digit_onehot_main}
\end{equation}
For each wavelet $\psi_i$ at scale $s_j$, we sample $\psi_{i,s_j,0}(t)$ on the same grid to obtain a discrete vector $\boldsymbol{\psi}_{i,s_j}\in\mathbb{R}^{B}$, and define the digit wavelet coefficient as
\begin{equation}
W_{\psi_i,s_j}(d) := \langle f_d, \boldsymbol{\psi}_{i,s_j} \rangle = (\boldsymbol{\psi}_{i,s_j})_{q(d)}.
\label{eq:digit_wavelet_coeff_main}
\end{equation}
This definition is consistent with the continuous formulation using an impulse and avoids the trivial zero coefficients produced by projecting constants onto zero-mean wavelets.

\paragraph{Digit embedding and dimension matching.}
We concatenate multi-wavelet, multi-scale coefficients into a feature vector
\begin{equation}
\phi(d) = \mathrm{vec}\left(\left[W_{\psi_i,s_j}(d)\right]_{i=1..k,\, j=1..l}\right) \in \mathbb{R}^{kl}.
\label{eq:digit_feature_vec_main}
\end{equation}
To interface with an LLM of embedding dimension $D$, we map $\phi(d)$ to a token embedding $E(d)\in\mathbb{R}^{D}$ via a fixed mapping $g(\cdot)$, which can be zero-padding when $kl\le D$ or a lightweight linear projection when $kl\ne D$:
\begin{equation}
E(d) = g(\phi(d)) \in \mathbb{R}^{D}.
\label{eq:digit_embed_map_main}
\end{equation}
Since there are only ten digits, $E(0),\ldots,E(9)$ can be precomputed and cached as a small embedding table.

\paragraph{TempoWave for a real number and injection into LLMs.}
Let $x$ be formatted into a fixed-precision digit sequence $(d_1,\ldots,d_{N_{dig}})$ with $N_{dig}=m_{prec}+n_{prec}$. TempoWave represents $x$ as a sequence of digit token embeddings
\begin{equation}
\mathrm{TempoWave}(x) = \left[E(d_1), E(d_2), \ldots, E(d_{N_{dig}})\right].
\label{eq:TempoWave_number_seq_main}
\end{equation}
In the input prompt, each digit token is embedded by $E(d_i)$, while other tokens use the original LLM embedding lookup. Standard positional encodings are applied as usual.

\paragraph{Summary on generation algorithm.}
 Given $x$, we (1) extract digits according to $(m_{prec},n_{prec})$, (2) compute $\phi(d)$ by evaluating wavelet samples at $q(d)$ for each $(\psi_i,s_j)$, (3) obtain $E(d)$ via $g(\cdot)$, and (4) assemble the digit-embedding sequence $\mathrm{TempoWave}(x)$.

\subsection{Representation Faithfulness in LLMs}
\label{sec:TempoWave_theory}

TempoWave is designed to provide a faithful and stable numeric interface for large language models by explicitly accounting for both the discrete nature of digits and the architectural properties of Transformers.
A central challenge in this setting is the pervasive use of normalization layers, such as LayerNorm and RMSNorm, which rescale and re-center token embeddings at every layer.
When numerical information is encoded primarily through absolute magnitudes, such normalization can severely distort or even collapse numerical distinctions, especially under deep stacking and autoregressive decoding.

The key design principle of TempoWave is to encode digits through \emph{structured multi-scale patterns} rather than raw scalar values.
Each digit is mapped to a vector of wavelet coefficients across multiple wavelet families and scales, capturing characteristic geometric patterns in the coefficient space.
By concatenating these coefficients and applying a fixed dimension-alignment mapping, TempoWave constructs digit embeddings whose identity is determined by relative patterns instead of absolute scale.
As a result, subsequent normalization operations mainly act as global affine transformations and do not destroy the structural differences between digits.

From a representational standpoint, this construction induces a \emph{finite digit codebook} in the embedding space.
Because the digit set is finite, injectivity of this codebook implies the existence of a positive separation margin between different digits, which guarantees robust nearest-neighbor recoverability under small perturbations.
This property underlies \textit{digit recoverability} and, by extension, \textit{numeracy preservation} under fixed precision, since each digit can be recovered independently from its embedding.

The use of multiple wavelets and multiple scales further enhances this separation.
Concatenating coefficients across wavelet-scale pairs cannot decrease pairwise distances between digit embeddings and typically increases them, thereby improving or maintaining the separation margin.
This explains why TempoWave exhibits {enhanced discriminability} compared to single-scale or single-frequency numeric encodings, as formally analyzed in the appendix.

Crucially, we also analyze how the induced digit codebook behaves under common normalization layers in Transformers.
We show that LayerNorm and RMSNorm can only collapse two embeddings under highly restricted affine conditions.
As long as the normalized digit codebook remains injective, digit identities remain uniquely recoverable after normalization.
Empirically, the multi-wavelet construction yields well-separated digit embeddings that remain distinct throughout the LLM.

\begin{table*}[t]
\setlength\tabcolsep{3pt}
\centering
\begin{footnotesize}

\resizebox{1\textwidth}{!}{
\begin{tabular}{lcccccccccc}
\toprule
& \multicolumn{2}{c}{\texttt{AUL}} & \multicolumn{2}{c}{\texttt{BIT}} & \multicolumn{2}{c}{\texttt{MSPG}} & \multicolumn{2}{c}{\texttt{PTF}} & \multicolumn{2}{c}{\texttt{LEU}} \\
\cmidrule(lr){2-3} \cmidrule(lr){4-5} \cmidrule(lr){6-7} \cmidrule(lr){8-9} \cmidrule(lr){10-11}
\textbf{Model} & RMSE & MAE & RMSE & MAE & RMSE & MAE & RMSE & MAE & RMSE & MAE \\
\midrule

\multicolumn{11}{c}{\textbf{\textit{Time Series Forecasting Model}}} \\
\midrule
DLinear & 0.5955 & 0.4977 & 1.5160 & 1.4051 & 0.4287 & 0.2901 & 0.4632 & 0.3268 & 0.6422 & 0.5355 \\
Autoformer & 0.8336 & 0.6828 & 1.4385 & 1.2953 & 0.4501 & 0.2998 & 0.3706 & 0.2288 & 0.6805 & 0.5937 \\
TimesNet & 0.5431 & 0.4595 & 1.2870 & 1.1874 & 0.3821 & 0.2792 & 0.3437 & 0.2238 & 0.3895 & 0.3303 \\
PatchTST & 0.4885 & 0.4086 & 1.2723 & 1.1684 & 0.4551 & 0.3098 & 0.3755 & 0.2492 & 0.4536 & 0.3634 \\
iTransformer & 0.6054 & 0.4948 & 1.3875 & 1.2675 & 0.3970 & 0.2711 & 0.3363 & 0.2234 & 0.6049 & 0.4948 \\
N-BEATS & 0.7158 & 0.5874 & 1.3603 & 1.1843 & 0.4305 & 0.2891 & 0.3713 & 0.2306 & 0.7480 & 0.5851 \\
\midrule

\multicolumn{11}{c}{\textbf{\textit{Time Series Foundation Model}}} \\
\midrule
TimeLLM & 0.7660 & 0.6012 & 1.3127 & 1.1538 & 0.4378 & 0.3267 & 0.5744 & 0.4730 & 0.4925 & 0.3701 \\
Chronos & 0.5970 & 0.4892 & 1.3348 & 1.1355 & 0.4103 & 0.2393 & 0.3606 & 0.2828 & 0.4184 & 0.2441 \\
Moirai & 0.5806 & 0.4854 & 1.3924 & 0.9823 & 0.3760 & 0.2368 & 0.2141 & 0.1681 & 0.3271 & 0.2267 \\
ChatTime-Chat & \underline{0.3639} & 0.3050 & \underline{0.8357} & \underline{0.7421} & 0.3123 & \textbf{0.1917} & \textbf{0.1610} & \underline{0.1276} & \textbf{0.1557} & \underline{0.1253} \\
\midrule

\multicolumn{11}{c}{\textbf{\textit{Fourier Embedding + LLM}}} \\
\midrule
FoNE-Qwen2.5-1.5b-instruct & 0.3649 & \underline{0.3045} & 1.7141 & 1.5202 & \underline{0.3006} & 0.2660 & 0.2915 & 0.2495 & 0.2065 & 0.1378 \\
\midrule

\multicolumn{11}{c}{\textbf{\textit{TempoWAVE Embedding + LLM}}} \\
\midrule
MWNE-Qwen2.5-1.5b-instruct & \textbf{0.3391} & \textbf{0.2739} & \textbf{0.7979} & \textbf{0.6978} & \textbf{0.2950} & \underline{0.1956} & \underline{0.1706} & \textbf{0.1212} & \underline{0.1681} & \textbf{0.1095} \\
\midrule

\multicolumn{11}{l}{\textbf{Relative $\downarrow$ Improvement over Previous Best (\%)}} \\
\midrule
\textbf{TempoWAVE vs Previous SOTA} & \textbf{7.3}\% & \textbf{11.2}\% & \textbf{4.7}\% & \textbf{6.3}\% & \textbf{1.9}\% & -2.0\% & -5.9\% & \textbf{5.3}\% & -7.9\% & \textbf{14.4}\% \\
\bottomrule
\end{tabular}
}
\caption{
\textbf{Forecasting performance (RMSE/MAE) across five context-enriched datasets.}
TempoWAVE achieves new SOTA on 7/10 metrics and ranks second on the remaining three.
Best values are \textbf{bolded}, second-best are \underline{underlined}.
The last row reports the relative $\downarrow$ improvement of MWNE over the previous best method for each metric.
}
\label{tab:ForecastPerformanceNew}
\end{footnotesize}
\vskip -0.15in
\end{table*}

%\ref{tab:ForecastPerformanceNew}

\begin{table*}[t]
\setlength\tabcolsep{3pt}
\centering
\begin{footnotesize}
\label{tab:context_abletion}
\resizebox{1\textwidth}{!}{
\begin{tabular}{lcccccccccc}
\toprule
& \multicolumn{2}{c}{\texttt{AUL}} & \multicolumn{2}{c}{\texttt{BIT}} & \multicolumn{2}{c}{\texttt{MSPG}} & \multicolumn{2}{c}{\texttt{PTF}} & \multicolumn{2}{c}{\texttt{LEU}} \\
\cmidrule(lr){2-3} \cmidrule(lr){4-5} \cmidrule(lr){6-7} \cmidrule(lr){8-9} \cmidrule(lr){10-11}
\textbf{Setting} & RMSE & MAE & RMSE & MAE & RMSE & MAE & RMSE & MAE & RMSE & MAE \\
\midrule

w/o context & 0.3809 & 0.3149 & 0.8356 & 0.7261 & 0.3218 & \underline{0.1917} & 0.2981 & 0.2391 & 0.2099 & 0.1613 \\
w/o Catch22 & 0.3759 & 0.3121 & 0.8205 & 0.7102 & \underline{0.3023} & 0.2057 & 0.2647 & 0.2119 & 0.1941 & \underline{0.1266} \\
w/o Situational context & \underline{0.3482} & \underline{0.2823} & \underline{0.8044} & \underline{0.7007} & 0.3123 & \textbf{0.1901} & \underline{0.1795} & \underline{0.1272} & \underline{0.1843} & 0.1284 \\
Full context & \textbf{0.3391} & \textbf{0.2739} & \textbf{0.7979} & \textbf{0.6978} & \textbf{0.2950} & 0.1956 & \textbf{0.1706} & \textbf{0.1212} & \textbf{0.1681} & \textbf{0.1095} \\

\bottomrule
\end{tabular}
}
\caption{Ablation Study: Forecasting performance (RMSE/MAE) across datasets under different context settings.}
\label{tab:Abl}
\end{footnotesize}
%\vskip -0.15in
\end{table*}

\begin{tcolorbox}[
    colback=blue!5!white,
    colframe=blue!75!black,
    title=\textbf{Full-context prompt example},
    fonttitle=\bfseries
]

\textbf{Input:} Historical load time series (48 half-hour points), e.g.,  
``..., -0.3849, -0.4859, -0.6162, -0.7185, ...''

\textbf{Context:}  
Region: VIC; Dates: 2021-05-12 $\rightarrow$ 2021-05-13 (weekday, non-holiday);  
Resolution: 30 minutes; Horizon: next 24 hours (48 points);  
Auxiliary features: daily weather statistics (temperature, humidity, pressure).

\textbf{Instruction:}  
Predict the next 48 load values using the time series and contextual information.  
Similarity is assessed using Catch22 statistical descriptors to preserve
autocorrelation, periodicity, and fluctuation patterns.

\textbf{Output:} Predicted sequence only, e.g.,  
“..., 0.3918, 0.3817, 0.4148, 0.4327, 0.4201, ...”

\end{tcolorbox}

\section{Experiments}
\label{sec:experiments}

\subsection{Experimental Setup}
\label{app:dataset_desc}

\paragraph{Datasets.}
We evaluate TempoWave on context-enriched forecasting benchmarks where each time series segment is paired with additional textual or event-based context.
First, we use the CGTSF dataset released via Hugging Face Datasets~\cite{liu2024chattime}, which contains three collections:
MSPG (solar power generation from 27 sites in Melbourne, 2021--2022, 15-minute frequency),
LEU (electricity usage from 16 London households, 2012--2013, 30-minute frequency),
and PTF (traffic flow from 32 Paris detectors in Paris, 2012, hourly frequency).
Each example includes a historical numerical window and associated context such as background descriptions, weather information (from Open-Meteo), date and holiday indicators, and curated news text when available.
We follow the official data splits and preprocessing protocol provided by the dataset source.

We additionally use the context-aware forecasting datasets from~\cite{wang2024newsforecast}, including Australia (AUL) and Bitcoin (BIT), which pair time series with relevant news articles.
For AUL and BIT, we follow the original preprocessing, normalization, and train/validation/test splits to ensure comparability with prior work.

\paragraph{Task formulation.}
Given a historical window of observations and its associated context, the model predicts the next $k$ future values.
We adopt a generative formulation: each numeric value is rendered into a fixed-precision string (e.g., \texttt{V.FFFF}) and the model generates future values as token sequences.
During fine-tuning, we minimize the standard cross-entropy loss over next-token prediction.

\paragraph{Decoding and numeric parsing.}
At inference time, generated token sequences are converted back to real values by parsing the fixed-precision numeric strings. An example of the full-context prompt is detailed in the accompanying text box.
If a generated output violates the numeric format (e.g., missing digits or containing non-numeric tokens), we apply a deterministic fallback parsing rule; if parsing still fails, the prediction for that step is treated as invalid and is counted in the evaluation according to the protocol.
All formatting and parsing rules are fixed across methods to ensure a fair comparison.

\paragraph{Evaluation metrics.}
Forecasting accuracy is measured using Mean Absolute Error (MAE) and Root Mean Squared Error (RMSE) across multiple prediction horizons.
Metrics are computed after inverting dataset-specific normalization when applicable, following the evaluation protocol of the corresponding benchmarks.

\paragraph{Baselines and fairness.}
We compare TempoWave-enhanced LLMs against a comprehensive set of baselines.
\textbf{(i) LLM-based baselines.} These methods use the same prompt templates and contextual inputs as TempoWave and differ only in the numeric interface, including standard tokenization and alternative input adaptation strategies.
\textbf{(ii) Time-series-specific baselines.} We also report results from established forecasting models that primarily operate on numerical history, including DLinear~\cite{zeng2023transformers}, N-BEATS~\cite{oreshkin2020nbeats}, Informer~\cite{zhou2021informer}, Autoformer~\cite{wu2021autoformer}, and TimesNet~\cite{timesnet}, as well as large-scale time series foundation models such as Chronos~\cite{ansari2024chronos} and Moirai~\cite{woo2024moirai}.
\textbf{(iii) Context-aware and embedding-interface baselines.} We include ChatTime~\cite{liu2024chattime} as a representative multimodal LLM system for forecasting with text context, and FoNE~\cite{zhou2025fone} as an alternative numerical embedding interface.

\subsection{Main Results}
\label{sec:main_results}

Table~\ref{tab:ForecastPerformanceNew} summarizes forecasting performance on five context-enriched benchmarks spanning news-driven series (\texttt{AUL}, \texttt{BIT}) and sensor or infrastructure series (\texttt{MSPG}, \texttt{PTF}, \texttt{LEU}).
Overall, TempoWave establishes a new state of the art on {7 out of 10} reported metrics and achieves {top-2} performance on all metrics.
Relative to the previous best method per metric (last row of Table~\ref{tab:ForecastPerformanceNew}), TempoWave yields an average {7.0\%} relative improvement on MAE across datasets, with the largest gains on \texttt{LEU} ({14.4\%}) and \texttt{AUL} ({11.2\%}).

\paragraph{Dataset-wise improvements and robustness.}
TempoWave delivers the most consistent gains on news-driven datasets.
On \texttt{AUL}, TempoWave improves both RMSE and MAE over the previous best by {7.3\%} and {11.2\%}, respectively.
On \texttt{BIT}, TempoWave achieves {4.7\%} (RMSE) and {6.3\%} (MAE) improvements over the previous best.
On \texttt{MSPG}, TempoWave achieves the best RMSE ({1.9\%} improvement) while remaining close to the best MAE (within {2.0\%} relative to the previous best).
For \texttt{PTF} and \texttt{LEU}, TempoWave attains the best MAE ({5.3\%} and {14.4\%} improvements), and achieves the second-best RMSE with small absolute gaps to the best baseline (0.0096 on \texttt{PTF}, 0.0124 on \texttt{LEU}).

\paragraph{MAE improves more consistently than RMSE.}
A recurring pattern is that TempoWave improves MAE more consistently than RMSE.
In particular, TempoWave reduces MAE on 4/5 datasets, while RMSE improvements are observed on 3/5 datasets.
This discrepancy is expected because RMSE emphasizes rare large deviations, whereas MAE better reflects typical per-step errors.

\paragraph{Comparison to numeric interfaces and time-series baselines.}
Compared with the Fourier-based numeric interface (FoNE) using the same LLM backbone, TempoWave is substantially more robust across domains.
The advantage is most prominent on \texttt{BIT}, where TempoWave reduces RMSE from 1.71 to 0.80 and MAE from 1.52 to 0.70, indicating improved generalization under highly non-stationary and event-driven dynamics.
Finally, TempoWave-enhanced LLMs outperform classic time-series forecasting models across all datasets in Table~\ref{tab:ForecastPerformanceNew}, highlighting the benefit of combining external context with a numerically grounded embedding interface.

\section{Analysis}

\subsection{Ablation Study on Contextual Information}
\label{sec:ablation_context}

To better understand how TempoWave interacts with different forms of contextual information in time series forecasting, we conduct a systematic ablation study over four context configurations, as summarized in Table~\ref{tab:Abl}.
Across five diverse datasets (\texttt{AUL}, \texttt{BIT}, \texttt{MSPG}, \texttt{PTF}, and \texttt{LEU}), we progressively remove components from the full context setting to isolate their individual and combined effects.

\paragraph{Overall impact of contextual information.}
The results show a clear and consistent trend: incorporating richer contextual information leads to improved forecasting performance across all datasets.
The full context setting achieves the best RMSE on all five datasets and the best MAE on four out of five datasets.
In contrast, removing all contextual information results in the weakest performance, indicating that TempoWave alone, while effective, benefits substantially from complementary contextual signals.
This trend is particularly pronounced on news-driven datasets such as \texttt{AUL} and \texttt{BIT}, where RMSE improves from 0.3809 to 0.3391 on \texttt{AUL} and from 0.8356 to 0.7979 on \texttt{BIT} when moving from no context to full context.

\paragraph{Contribution of different context components.}
Comparing partial ablations reveals that different types of context contribute in distinct and complementary ways.
Removing Catch22 features (\emph{w/o Catch22}) leads to noticeable degradation across most datasets, suggesting that statistical descriptors capturing autocorrelation, periodicity, and distributional properties provide strong global signals for forecasting.
Indeed, the \emph{w/o Catch22} setting consistently underperforms the full context configuration.
Conversely, removing situational context (\emph{w/o situational context}) primarily affects datasets with strong external dependencies, such as \texttt{AUL} and \texttt{BIT}, where date and domain-related information play a more prominent role.

\paragraph{Dataset-specific behavior.}
The relative importance of context components varies across datasets.
For infrastructure and sensor-driven datasets (\texttt{MSPG}, \texttt{PTF}, and \texttt{LEU}), Catch22 features alone already provide strong performance, in some cases matching or approaching the full context results.
For example, on \texttt{MSPG}, the \emph{w/o situational context} setting achieves the best MAE (0.1901), indicating that short-term statistical regularities dominate forecasting performance.
In contrast, on \texttt{AUL} and \texttt{BIT}, which are influenced by external events and news, the full context setting yields the largest gains, highlighting the importance of integrating situational and textual information with TempoWave.

\begin{figure}[t]
    \centering
    \includegraphics[width=0.9\linewidth]{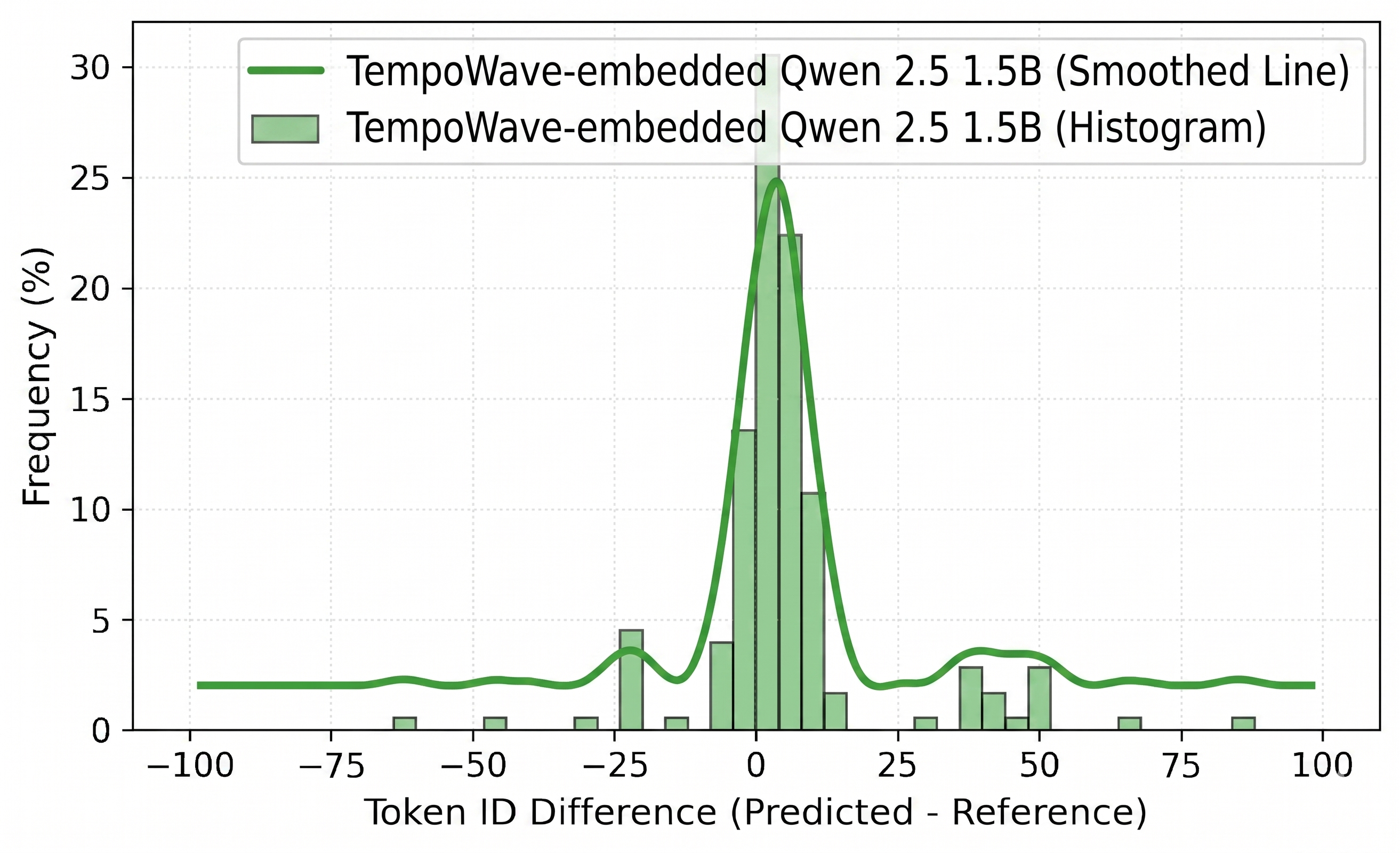}
    \caption{
Token ID difference distribution between predicted tokens and their reference counterparts for \textbf{TempoWave-embedded Qwen 2.5 1.5B model}, under the top-10 prediction setting. The histogram illustrates raw frequency, while the smoothed curve highlights the overall trend. The sharp concentration around zero indicates strong local proximity in token prediction.
}
    \label{fig:TempoWave-distribution}
\end{figure}

\begin{figure}[t]
    \centering
    \includegraphics[width=0.9\linewidth]{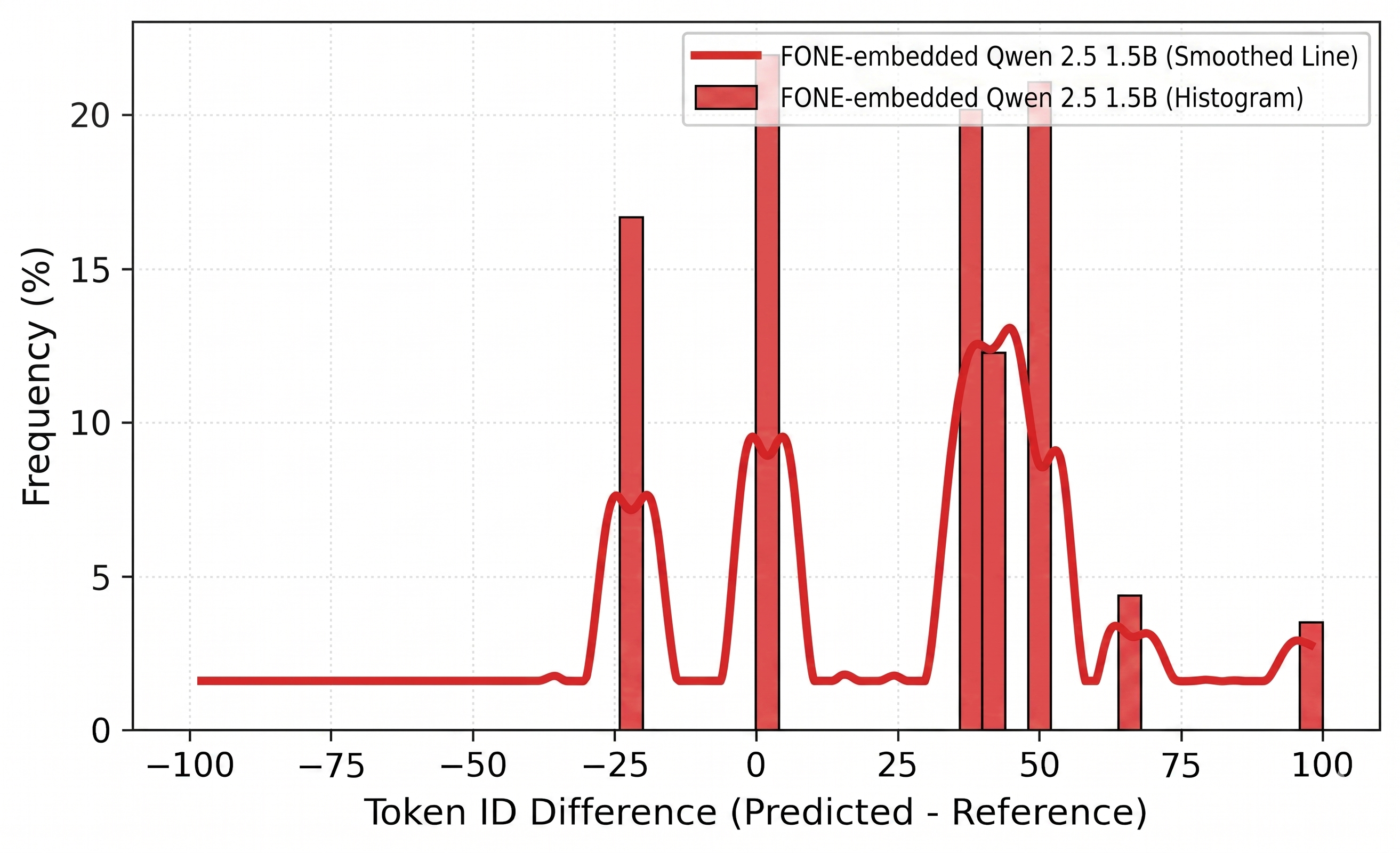}
    \caption{
Token ID difference distribution between predicted tokens and their reference counterparts for \textbf{FoNE-embedded Qwen 2.5 1.5B model} (baseline), under the top-10 prediction setting. The histogram illustrates raw frequency, while the smoothed curve highlights the overall trend. The sharp concentration around zero indicates strong local proximity in token prediction.
}
    \label{fig:fone-distribution}
\end{figure}

\begin{figure}[t]
    \centering
    \includegraphics[width=0.9\linewidth]{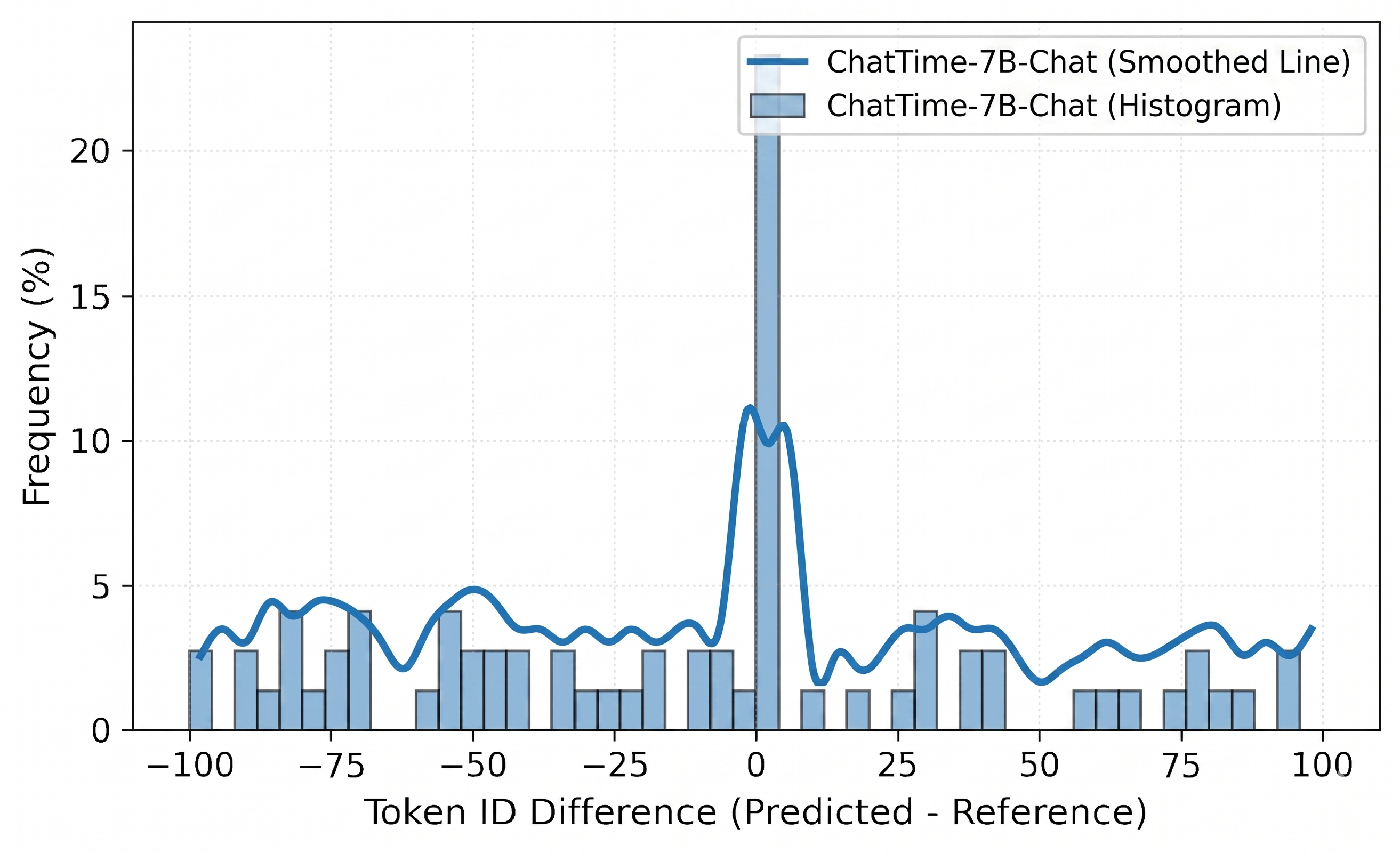}
    \caption{
Token ID difference distribution between predicted tokens and their reference counterparts for \textbf{ChatTime-7B-Chat model} (baseline), under the top-10 prediction setting. The histogram illustrates raw frequency, while the smoothed curve highlights the overall trend. The sharp concentration around zero indicates strong local proximity in token prediction.
}
    \label{fig:chattime-distribution}
\end{figure}

\subsection{Embedding Alignment via Next Token Proximity}
To evaluate the semantic and structural alignment of different embedding strategies, we analyze the distribution of token ID proximity between the model’s predicted next token and the immediately preceding token in the input prompt. This probing task is particularly informative in our setting, where tokens represent numerical values derived from time series data. A well-structured embedding should induce a smooth, symmetric distribution reflecting temporal continuity. Our method, as shown in Figure~\ref{fig:TempoWave-distribution}, exhibits a clear unimodal, approximately Gaussian distribution centered around zero, indicating that the model learns to predict numerically coherent tokens aligned with the underlying time series dynamics. In contrast, the FoNE baseline in Figure~\ref{fig:fone-distribution}, evaluated with the same backbone but without an explicit numeric inductive bias, exhibits a flatter and more irregular distribution, indicating weaker alignment with underlying numeric trends. More notably, a standard pretrained baseline without our embedding augmentation in Figure~\ref{fig:chattime-distribution} exhibits a sharp, anomalous spike in one bin, revealing a tendency to overfit by repeatedly predicting a fixed token, regardless of local context. These results underscore the effectiveness of our embedding approach in capturing latent numerical semantics and encoding smooth transitions that mirror real-world time series behavior.

\section{Conclusion}

While directly applying large language models (LLMs) to time series analysis remains challenging due to the mismatch between continuous values and discrete token interfaces, the potential payoff is substantial. In this paper, we proposed {Multi-Wavelet Number Embedding (TempoWave)}, a numerically grounded embedding interface that leverages multi-resolution wavelet features to bridge the numerical--textual modality gap for time series forecasting. Extensive experiments on five diverse benchmarks show that TempoWave consistently improves LLM-based forecasters, outperforming strong specialized time series models and alternative numeric embedding approaches in most settings. Empirically, TempoWave is more robust under non-stationarity and extreme values, and exhibits favorable optimization behavior, including smoother training dynamics, resilience to digit-level perturbations, and stable interaction with common normalization layers. Ablation results further highlight that contextual information is complementary to TempoWave and contributes to the strongest overall performance. Together, these findings advance LLM-based forecasting by coupling LLMs' contextual reasoning with a more faithful numeric interface. A promising direction for future work is to investigate whether TempoWave also benefits non-contextual forecasting pipelines that rely on discretization or binning-based tokenization of time series values.

\newpage
\section*{Ethical Statement}

There are no ethical issues.

\section*{Acknowledgements}

This work is partially supported by the NSF Award \#2425919, and NSF Award \#2413417. The funding from these sources has been a cornerstone in enabling us to bring our project to fruition. We are also deeply grateful to the anonymous reviewers for their rigorous review process. Their detailed comments and constructive suggestions have significantly contributed to the improvement of this paper.

%% The file named.bst is a bibliography style file for BibTeX 0.99c
\bibliographystyle{named}
\bibliography{ijcai26}

\end{document}